\documentclass[letterpaper, 10 pt, conference]{ieeeconf}
\pdfoutput=1
\IEEEoverridecommandlockouts
\UseRawInputEncoding

\usepackage[T1]{fontenc}
\usepackage{lmodern}

\usepackage{afterpage}
\usepackage[utf8]{inputenc}
\usepackage{times}
\usepackage{graphicx}
\usepackage{amsmath,amssymb,bm}
\usepackage{algorithm}
\usepackage{algorithmic}
\usepackage[hyphens]{url}
\usepackage{xcolor}
\usepackage{booktabs}
\usepackage{multirow}
\usepackage{xspace}
\let\labelindent\relax
\usepackage{enumitem}
\usepackage{microtype}
\usepackage{cite}
\usepackage[colorlinks=true,linkcolor=black,citecolor=black,urlcolor=black]{hyperref}

\title{\LARGE \bf
Reinforcement Fine-Tuning of Flow-Matching Policies for Vision-Language-Action Models
}

\author{%
Mingyang Lyu\textsuperscript{a,b,e,1},
Yinqian Sun\textsuperscript{a,b,d,f,1},
Erliang Lin\textsuperscript{a},
Huangrui Li\textsuperscript{d}, \\ 
Ruolin Chen\textsuperscript{a,b,e},
Feifei Zhao\textsuperscript{a,b,d,f,2},
Yi Zeng\textsuperscript{a,b,c,d,e,f,2}%
\thanks{Author affiliations: \textsuperscript{a} Brain-inspired Cognitive AI Lab, Institute of Automation, Chinese Academy of Sciences, Beijing, China.; \textsuperscript{b} Beijing Institute of AI Safety and Governance, China.;
\textsuperscript{c}State Key Laboratory of Brain Cognition and Brain-inspired Intelligence Technology;
\textsuperscript{d} Beijing Key Laboratory of Safe AI and Superalignment, China.; \textsuperscript{e} University of Chinese Academy of Sciences (UCAS), Beijing, China.; \textsuperscript{f} Long-term AI,Beijing,China.}%
\thanks{\textsuperscript{1} Co-first authors. \textsuperscript{2} Co-corresponding authors.
Correspondence: \texttt{zhaofeifei2014@ia.ac.cn} (Feifei Zhao), \texttt{yi.zeng@ia.ac.cn} (Yi Zeng).
Additional contact: \texttt{lvmingyang2024@ia.ac.cn} (Mingyang Lyu).}%
}

\begin{document}
\maketitle
\thispagestyle{empty}
\pagestyle{empty}

\begin{abstract}

Vision-Language-Action (VLA) models such as OpenVLA, Octo, and $\pi_0$ have shown strong generalization by leveraging large-scale demonstrations, yet their performance is still fundamentally constrained by the quality and coverage of supervised data. 
Reinforcement learning (RL) provides a promising path for improving and fine-tuning VLAs through online interaction. However, conventional policy gradient methods are computationally infeasible in the context of flow-matching based models due to the intractability of the importance sampling process, which requires explicit computation of policy ratios. To overcome this limitation, we propose Flow Policy Optimization (FPO) algorithm, which reformulates importance sampling by leveraging per-sample changes in the conditional flow-matching objective. 
Furthermore, FPO achieves stable and scalable online reinforcement fine-tuning of the $\pi_0$ model by integrating structure-aware credit assignment to enhance gradient efficiency, clipped surrogate objectives to stabilize optimization, multi-step latent exploration to encourage diverse policy updates, and a Q-ensemble mechanism to provide robust value estimation. We evaluate FPO on the LIBERO benchmark and the ALOHA simulation task against supervised, preference-aligned, diffusion-based, autoregressive online RL, and $\pi_0$-FAST baselines, observing consistent improvements over the imitation prior and strong alternatives with stable learning under sparse rewards. In addition, ablation studies and analyses of the latent space dynamics further highlight the contributions of individual components within FPO, validating the effectiveness of the proposed computational modules and the stable convergence of the conditional flow-matching objective during online RL.

\end{abstract}

\section{INTRODUCTION}

The pursuit of generalist robots capable of executing a diverse array of physical tasks has advanced significantly with the emergence of Vision-Language-Action (VLA) models. Recent architectures, such as OpenVLA \cite{OpenVLA2024} and Octo \cite{Octo2024}, have demonstrated that policies pre-trained on large-scale datasets of human demonstrations can acquire broad semantic understanding and effectively execute a wide spectrum of instructions. Notably, the $\pi_0$ \cite{pi0_2024} model implements action generation via a flow-matching technique \cite{Lipman2022FlowMatching,Tong2025CFM}. This method confers a unique advantage: it enables the generation of smooth, temporally coherent, high-frequency action segments, which is essential for achieving dexterous and long-horizon manipulation tasks that require more than isolated, single-step action predictions.

Drawing inspiration from the remarkable progress of Reinforcement Learning (RL) in enhancing Large Language Models (LLMs) beyond their supervised fine-tuning (SFT) performance \cite{RLHF_LLM1, RLHF_LLM2, RLHF_LLM3}, there is a growing trend to apply RL for post-training of embodied VLA models. Approaches such as online RL for auto-regressive VLAs \cite{VLA_RL2025}, iterative RL+SL stabilization for large VLAs \cite{iReVLA2025}, and policy-gradient fine-tuning of diffusion/flow-matching policies\cite{DPPO2024,ReinFlow2025} have demonstrated that robotic agents can leverage online interaction to refine skills and discover strategies superior to those available in initial imitation datasets. This paradigm allows agents to overcome the inherent limitations of offline demonstration data quality and coverage, pushing performance beyond the imitation ceiling.

However, a core technical incompatibility arises when applying conventional RL techniques to flow-matching-based VLA models such as $\pi_0$ \cite{pi0_2024}. Commonly used policy gradient methods for reinforcement fine-tuning of VLA models, such as PPO \cite{Schulman2017PPO} and TRPO \cite{Schulman2015TRPO}, require importance sampling, i.e., the explicit computation of policy ratios. For flow-matching models, this computation is analytically intractable. It necessitates solving an underlying ordinary differential equation \cite{Chen2018NeuralODE} and integrating a computationally prohibitive Jacobian trace term along the generation path \cite{Grathwohl2018FFJORD}. This renders such methods computationally infeasible for the demands of online fine-tuning. While reward-weighted supervised learning approaches exist, they typically struggle with active exploration and the discovery of novel, out-of-distribution behaviors. These combined challenges have largely precluded the effective application of online RL for fine-tuning flow-matching generative policy-based VLA models.

In this paper, we introduce Flow Policy Optimization (FPO), a method designed to overcome the incompatibility between flow-matching policies and PPO-style updates by constructing a likelihood-free policy ratio based on per-sample changes in the conditional flow-matching objective. This formulation eliminates the need for explicit action likelihoods and ODE–Jacobian computations while preserving consistency with the policy’s generative structure. Furthermore, we provide structure-aware credit assignment in the latent space by leveraging the model’s training objective as a per-sample improvement signal. Combined with a clipped surrogate objective, multi-step latent exploration, and a Q-ensemble, FPO enables stable and efficient learning even in environments with sparse rewards and contact-rich dynamics. The proposed method successfully facilitates online reinforcement fine-tuning of the $\pi_0$ model, with its effectiveness and superiority empirically validated on the LIBERO benchmark and the ALOHA Transfer Cube task. In summary, the main contributions of this work are summarized as follows: 
\begin{itemize}
    \item We propose FPO, a practical policy optimization framework that bridges flow-matching policies and PPO-style updates by introducing a likelihood-free policy ratio derived from per-sample changes in the conditional flow-matching objective. This formulation avoids explicit density estimation and complex Jacobian computations, while retaining structural consistency with the generative policy.
    
    \item We develop an online reinforcement fine-tuning algorithm for the $\pi_0$ model by integrating structure-aware credit assignment in the latent space with key RL components including a clipped surrogate objective, multi-step latent exploration, and a Q-ensemble. This combination ensures stable and efficient learning in challenging environments with sparse rewards and contact-rich dynamics.

    \item Extensive experiments on the LIBERO benchmark and the ALOHA Transfer Cube task demonstrate the superior performance of $\pi_0$-FPO over six strong baselines such as OpenVLA, Octo, Diffusion Policy, GRAPE, and $\pi_0$-FAST, achieving an average success rate of 87.2$\%$ on LIBERO, 65.3$\%$ on LIBERO-Long, and more than $1.5\times$ the baseline success rate on ALOHA-sim. Ablation studies validate the contribution of each component, while qualitative and latent-space analyses highlight improved correction of recurrent failure modes.
\end{itemize}

\afterpage{%
  \begin{figure*}[!t]
    \centering
    \includegraphics[width=0.85\textwidth]{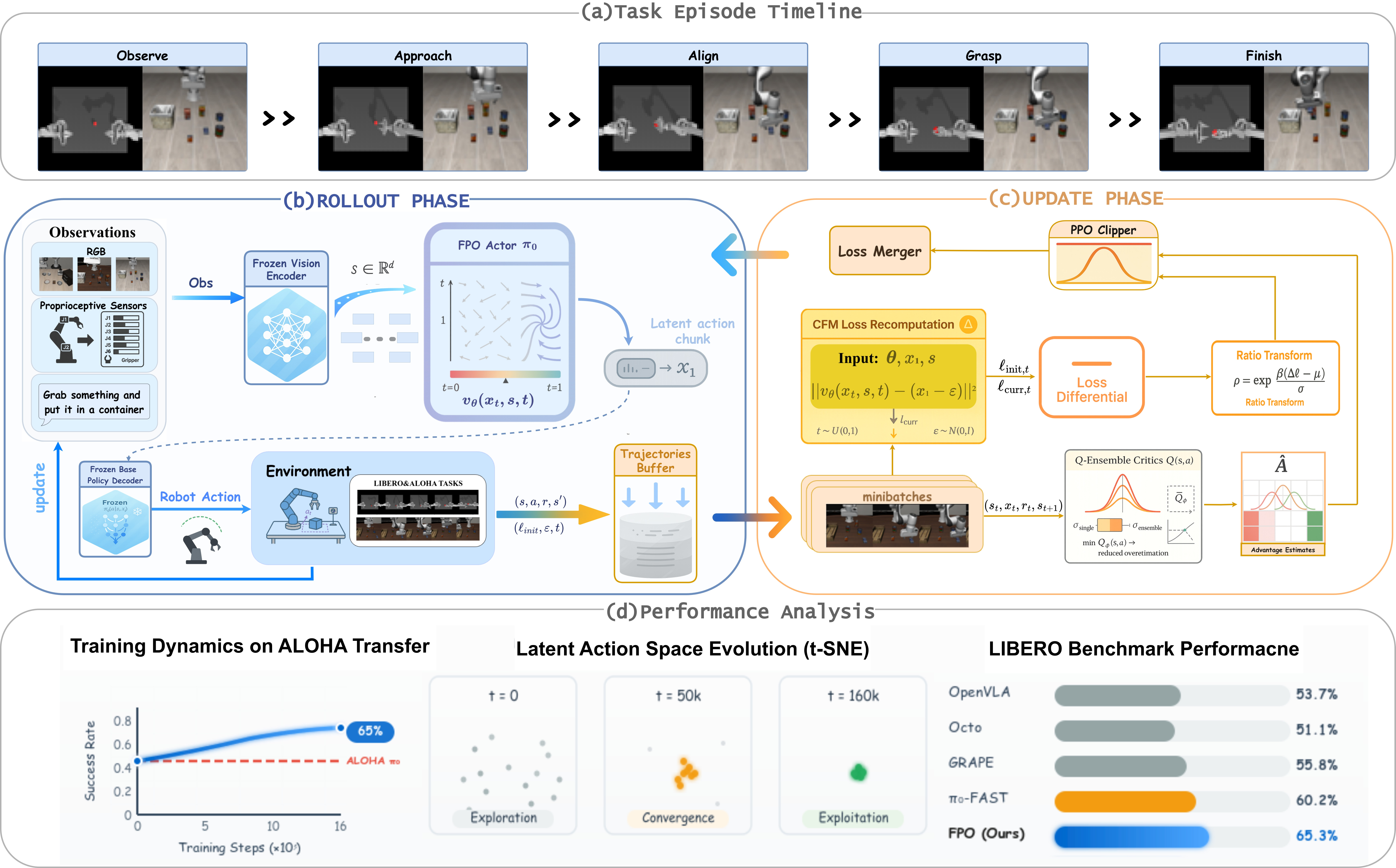}
    \caption{Overview of Flow Policy Optimization (FPO).
    (a) Task episode timeline.
    (b) Rollout phase: a frozen encoder produces state $s$, the actor $\pi_\theta$ outputs a latent chunk $x_1$, the frozen base policy $\pi_0$ decodes $(s,x_1)$ to control $a$, yielding $(r,s')$. We store the transition and cache the initial CFM loss in a near sliding-window trajectory buffer.
    (c) Update phase: Batch of trajectories are sampled, the CFM loss is recomputed to form a loss differential, which is mapped to a likelihood-free ratio. A Q-ensemble supplies advantages, and the actor is updated with a clipped surrogate. The updated policy feeds back to rollout.
    (d) Performance panels: example training curve, latent-space evolution (t-SNE), and LIBERO success rates.}
    \label{fig:overview}
  \end{figure*}
}
\section{RELATED WORK}
\subsection{Vision-Language-Action Models} 
VLA models are commonly trained via behavioral cloning on large-scale human demonstrations, coupling language and vision with end-to-end control~\cite{VLA_survey2025}. Early systems predominantly used autoregressive, token-based action decoders, recent policies replace discrete heads with diffusion- or flow-matching controllers~\cite{Ho2020DDPM,Song2021SDE}, capturing continuous, multi-modal action distributions and enabling smooth, high-frequency control for dexterous manipulation. Representative systems include Octo and $\pi_0$~\cite{Octo2024,pi0_2024}, 
whose generative action heads are grounded in the modeling literature and applied to visuomotor control~\cite{DiffusionPolicy2023,Lipman2022FlowMatching,Tong2025CFM}. In parallel, OpenVLA provides an open-source generalist policy that scales across embodiments and supports efficient fine-tuning~\cite{OpenVLA2024}, orthogonal advances include frequency-space action tokenization (FAST) for high-rate control~\cite{FAST2025}, BC-Z for zero-shot generalization from language goals~\cite{BCZ2022}, and preference alignment via GRAPE~\cite{GRAPE2024}, DPO~\cite{Rafailov2023DPO} and learning from human preferences~\cite{Christiano2017Prefs}. Together, these developments situate diffusion- and flow-matching decoders within a broader VLA trajectory~\cite{VLA_survey2025} and motivate online updates that improve beyond the imitation prior.
\subsection{Reinforcement Learning for VLA Policies}
Policy-gradient fine-tuning of generative policies faces two obstacles: intractable (or costly) likelihoods along generative trajectories and long-horizon credit assignment. For diffusion policies, PPO-style surrogates have been adapted to the denoising process with architecture-aware designs (DPPO)~\cite{DPPO2024}. For flow-matching policies, recent work follows two lines. One line performs reward-weighted supervised updates that avoid explicit likelihoods by biasing training toward high-return samples (RWFM and variants)~\cite{Pfrommer2025RWFM}. The other introduces stochastic relaxations or noise injection to enable sampling-based ratios and on-policy updates (Flow-GRPO~\cite{Liu2025FlowGRPO}, ReinFlow~\cite{ReinFlow2025}). Related gradient estimators for flow models have also been explored from a policy-gradient perspective~\cite{FlowRWFM2025}. In VLA settings with autoregressive heads, VLA-RL reports procedures for scaling online RL with trajectory-level optimization~\cite{VLA_RL2025}, while preference-based tuning connects to advances in RL for large models~\cite{RLHF_LLM1,RLHF_LLM2,RLHF_LLM3}. 
A persistent difficulty for flow-based actors is that exact policy ratios generally require solving probability-flow ODEs with Jacobian-trace terms~\cite{Song2021SDE}, rendering likelihoods and ratios expensive or intractable for online control~\cite{Chen2018NeuralODE,Grathwohl2018FFJORD}. We address this by constructing a likelihood-free ratio from per-sample changes in the conditional flow-matching objective and performing clipped PPO-style updates aligned with the model’s generative structure.

\section{METHOD}

The proposed FPO is an actor--critic framework that enables online fine-tuning of pretrained conditional flow-matching policies without requiring tractable action likelihoods. The central idea is to reformulate importance sampling by exploiting per-sample changes in the CFM objective as a structure-aligned signal, which is mapped to a likelihood-free ratio and used within a PPO-style clipped surrogate. To obtain stable and efficient updates, FPO integrates (i) structure-aware credit assignment in the action latent space, (ii) clipped surrogates for trust-region control, (iii) multi-step latent (Euler) exploration to produce smooth, temporally correlated perturbations, and (iv) a critic ensemble that supplies robust value estimates. 
Training alternates between \emph{rollout} and \emph{update} (Fig.~\ref{fig:overview}b,c, Algorithm~\ref{alg:fpo}): rollout logs transitions and per-sample CFM losses into a small sliding-window buffer; updates recompute losses under the current actor, map their differences to a clipped surrogate, and train the critic ensemble; the updated actor is then used for the next rollout. Formal details appear in Sec.~\ref{sec:fpo_pipeline_overview} and Sec.~\ref{sec:actor_critic_details}.

\begin{algorithm}[t]
\caption{Flow Policy Optimization (FPO)}
\label{alg:fpo}
\begin{algorithmic}[1]
\STATE \textbf{Input:} frozen base policy $\pi_{0}$, actor $\pi_{\theta}$, critic ensemble $\{Q_{\phi_i}\}_{i=1}^M$, target critics $\{Q_{\bar\phi_i}\}_{i=1}^M$, buffer $\mathcal{B}$
\FOR{each iteration}
    \STATE $\theta_{\text{old}} \leftarrow \theta$ \COMMENT{Freeze actor for loss caching}
    \STATE {\color{gray!50}// \textbf{Rollout phase}}
    \FOR{$t = 0 \dots T_{\text{rollout}}-1$}
        \STATE observe $s_t$
        \STATE sample latent $x_t \sim \pi_{\theta}(\cdot \mid s_t)$
        \STATE decode action $a_t \sim \pi_{0}(\cdot \mid s_t, x_t)$
        \STATE step env, obtain $(r_t, s_{t+1})$
        \STATE cache $\ell_{\text{init},t} \leftarrow \ell_{\text{cfm}}(x_t \mid s_t; \theta_{\text{old}})$
        \STATE push $(s_t, x_t, a_t, r_t, s_{t+1}, \ell_{\text{init},t})$ into $\mathcal{B}$ 
    \ENDFOR
    \STATE {\color{gray!50}// \textbf{Update phase}}
    \FOR{$k = 1 \dots K_{\text{update}}$}
        \STATE sample batch $\mathcal{M} \subset \mathcal{B}$
\STATE {\color{gray!50}// Critic update (Eqs.5,6)}
        \STATE for each $(s_t, x_t, r_t, s_{t+1}) \in \mathcal{M}$:
        \STATE \hspace{0.9em} sample $x'_{t+1} \sim \pi_{\theta}(\cdot \mid s_{t+1})$ and set
        \STATE \hspace{0.9em} $y_t \leftarrow r_t + \gamma \min_{i} Q_{\bar{\phi}_i}(s_{t+1}, x'_{t+1})$
        \STATE update $\{\phi_i\}$ by minimizing $\mathcal{L}_{\text{critic}}(\phi)$;
        \STATE Polyak update targets $Q_{\bar{\phi}_i}$
\STATE {\color{gray!50}// Actor update (Eqs.2,3,4)}
        \STATE compute $\Delta\ell_{\text{cfm},t} \leftarrow \ell_{\text{cfm}}(x_t \mid s_t; \theta_{\text{old}})\,\ell_{\text{cfm}}(x_t \mid s_t; \theta)$
        \STATE standardize $z_t \leftarrow \mathrm{standardize}(\Delta\ell_{\text{cfm},t})$
        \STATE map ratio proxy $\rho_t \leftarrow \exp(\beta z_t)$
        \STATE compute advantages $\hat{A}_t$ from the critic ensemble (e.g., GAE)
        \STATE update $\theta$ by minimizing $\mathcal{L}_{\text{actor}}(\theta)$
    \ENDFOR
\ENDFOR
\end{algorithmic}
\end{algorithm}
\subsection{FPO Pipeline and Problem Formulation}
\label{sec:fpo_pipeline_overview}
FPO steers a frozen base policy $\pi_{0}(a \mid s, x)$ with a flow-based actor $\pi_{\theta}(\cdot \mid s)$ operating in the action latent space. Let $x(u; s)$ denote the actor’s latent at flow time $u\!\in\![0,1]$ for state $s$, and write $x_t := x(1; s_t)$ for the latent produced at environment step $t$. A frozen encoder maps observations to $s_t\!\in\!\mathbb{R}^{d}$; the actor samples $x_t\!\in\!\mathbb{R}^{D}$; the base decodes $(s_t,x_t)$ to an action $a_t$; the environment yields reward $r_t$ and next state $s_{t+1}$ (Fig.~\ref{fig:overview}). The optimization objective is
\begin{equation}
  J(\theta) \;=\; \mathbb{E}\!\left[\sum_{t=0}^{T}\gamma^{t}\, r_t\right], \qquad \gamma \in (0,1).
  \label{eq:obj}
\end{equation}
Because the actor is a conditional flow model, $\log \pi_{\theta}(x_t \mid s_t)$ is generally intractable, precluding direct policy-ratio computation.

\paragraph{Rollout and Data Recording}
Training proceeds in alternating \emph{rollout} and \emph{update} phases (Fig.~\ref{fig:overview}b,c; Algorithm~\ref{alg:fpo}). During interaction, a frozen rollout copy $\theta_{\text{old}}$ is used to generate experience so that logged quantities remain consistent with the data-collecting policy. At each step, the encoder produces $s_t$, the actor samples a latent chunk $x_t$ (with optional short Euler perturbations in latent space for exploration), and the frozen base policy $\pi_0$ decodes $(s_t,x_t)$ to low-level control $a_t$ that is executed in the environment. The system records $(s_t, x_t, a_t, r_t, s_{t+1})$ and caches the per-sample CFM loss $\ell_{\text{cfm}}(x_t \mid s_t; \theta_{\text{old}})$ attached to the exact latent used for control. Transitions are stored in a small sliding-window \emph{trajectory buffer} that retains only recent rollouts. This design preserves the linkage between cached losses and their originating policy, and bounds the distributional gap between the data-collecting policy and the subsequently updated actor.

\paragraph{Update Cycle}
During updates, data are drawn from the trajectory buffer and the CFM loss is re-evaluated under the current actor $\theta$ on the same $(s_t,x_t)$ pairs. The resulting per-sample loss differential is converted, via batch-standardization and a monotone mapping, into a \emph{likelihood-free} ratio proxy that serves as the multiplicative factor in a PPO-style clipped surrogate. Advantages are supplied by a critic ensemble queried in latent space. Actor and critics are optimized for several SGD epochs per interaction batch while continuously evicting older trajectories to keep the training distribution close to recent behavior. Target networks for the critics are updated by Polyak averaging to stabilize bootstrapped targets. After the update cycle, parameters are synchronized by setting $\theta_{\text{old}}\!\leftarrow\!\theta$ before the next interaction phase. This schedule closes the interaction–update loop, maintains a tight coupling between the ratio proxy and the data-collecting policy, and yields steady improvement without requiring tractable action likelihoods.

\begin{figure*}[t]
    \centering
    \includegraphics[width=0.9\textwidth]{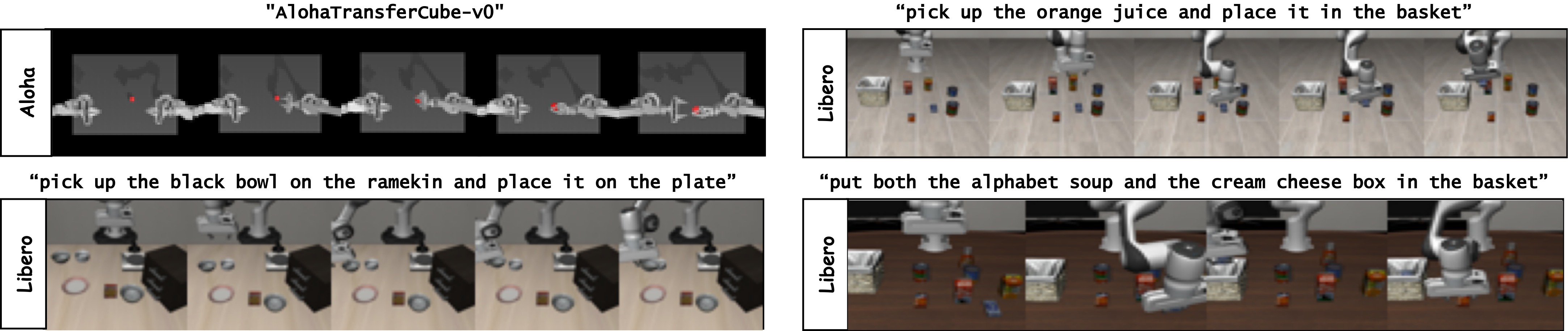}

    \caption{An overview of the challenging visuomotor control environments used in our evaluation: the bimanual ALOHA Transfer Cube task and several multi-object manipulation tasks from the LIBERO suite. These environments require a combination of long-horizon reasoning, precise control, and generalization across different objects and initial conditions.}
    \label{fig:environments}
\end{figure*}
\subsection{Structure-Aligned Policy Update and Training Components}
\label{sec:actor_critic_details}

\paragraph{Likelihood-Free Ratio from CFM Loss}
Because $\log\pi_{\theta}(x_t \mid s_t)$ is intractable for flow-based actors, FPO uses the actor’s CFM objective as the update signal. Let $\ell_{\text{cfm}}(x_t \mid s_t; \theta)$ denote the per-sample CFM loss~\cite{Lipman2022FlowMatching,Tong2025CFM}. For each stored pair $(s_t, x_t)$, the loss reduction is:
\begin{equation}
    \Delta\ell_{\text{cfm},t} \;=\; \ell_{\text{cfm}}(x_t \mid s_t; \theta_{\text{old}})\;-\;\ell_{\text{cfm}}(x_t \mid s_t; \theta)
    \label{eq:lossdrop}
\end{equation}
which measures improvement on the \emph{same} sample relative to the rollout actor. Under a mild local monotonicity assumption—that per-sample CFM loss decreases coincide with increases in the actor’s conditional density—we treat $\Delta\ell_{\text{cfm},t}$ as an order-preserving surrogate of the intractable importance ratio $\pi_\theta(x_t\!\mid\!s_t)/\pi_{\theta_{\text{old}}}(x_t\!\mid\!s_t)$. And the $\Delta\ell_{\text{cfm},t}$ is normalized with 
\begin{equation}
    z_t \;=\; \frac{\Delta\ell_{\text{cfm},t}-\mu_{\Delta}}{\sigma_{\Delta}}, \qquad
    \rho_t \;=\; \exp(\beta\, z_t)
    \label{eq:zscore_ratio}
\end{equation}
where $(\mu_{\Delta},\sigma_{\Delta})$ are mean and standard deviation as batch statistics and $\beta>0$ controls the sharpness of the mapping.

\paragraph{Clipped Surrogate and Actor Update}
The actor is optimized with a PPO-style clipped surrogate~\cite{Schulman2017PPO} using advantages $\hat A_t$:
\begin{equation}
    \mathcal{L}_{\text{actor}}(\theta)\;=\;-\mathbb{E}_t\Big[\min\!\big(\rho_t\,\hat{A}_t,\ \text{clip}(\rho_t,1-\epsilon,1+\epsilon)\,\hat{A}_t\big)\Big]
    \label{eq:actor_loss}
\end{equation}
where $\epsilon>0$ is the clipping parameter. This construction regulates update magnitude while preserving alignment with the actor’s generative structure.In practice, we standardize $\hat A_t$ within each minibatch and stop gradients through $\rho_t$ to reduce variance and avoid feedback instabilities.

\paragraph{Critic Ensemble and Advantage Estimation}
We employ an ensemble of action–value functions $\{Q_{\phi_i}(s,x)\}_{i=1}^M$ to reduce overestimation and stabilize advantage estimates. Target critics $Q_{\bar{\phi}_i}$ are updated by Polyak averaging once per gradient step. For a transition $(s_t,x_t,r_t,s_{t+1})$, the temporal-difference target is:
\begin{equation}
\resizebox{.85\hsize}{!}{$
y_t = r_t + \gamma \min_{i} Q_{\bar{\phi}_i}(s_{t+1}, x'_{t+1}),\quad
x'_{t+1}\sim\pi_\theta(\cdot\mid s_{t+1})
$}
\label{eq:tdtarget}
\end{equation}
where the operation of minimizing introduces a conservative target that empirically curbs optimistic bias. For terminal $s_{t+1}$ the bootstrap term is masked out. The critic loss is the squared TD error:
\begin{equation}
\mathcal{L}_{\text{critic}}(\phi)\;=\;\mathbb{E}\Big[\big(Q_{\phi}(s_t,x_t)-y_t\big)^2\Big]
\label{eq:critic_loss}
\end{equation}
Advantages are computed with generalized advantage estimation (GAE)~\cite{Schulman2015GAE}. The value baseline $V(s)$ is taken as a conservative estimate from the ensemble (the minimum across members). We reuse stored latents when available; otherwise a fresh latent is sampled from $\pi_\theta(\cdot\mid s)$.

\begin{table*}[t]
\centering
\caption{Overall success rate (SR, \%) and per-suite rank on LIBERO benchmarks.
Ranks are computed among baseline methods only (excluding $\pi_0$-FAST). Avg Rank is the mean of per-suite ranks for each baseline. Best in \textbf{bold}.}
\label{tab:libero}
\begin{tabular}{lcc|cc|cc|cc|cc}
\toprule
 & \multicolumn{2}{c}{\textbf{LIBERO-Spatial}}
 & \multicolumn{2}{c}{\textbf{LIBERO-Object}}
 & \multicolumn{2}{c}{\textbf{LIBERO-Goal}}
 & \multicolumn{2}{c}{\textbf{LIBERO-Long}}
 & \multicolumn{2}{c}{\textbf{Average}} \\
\textbf{Method} & \textbf{SR (\%)} & \textbf{Rank} & \textbf{SR (\%)} & \textbf{Rank} & \textbf{SR (\%)} & \textbf{Rank} & \textbf{SR (\%)} & \textbf{Rank} & \textbf{SR (\%)} & \textbf{Avg Rank} \\
\midrule
Diffusion Policy\cite{DiffusionPolicy2023}& 78.3 & 6 & 92.5 & 2 & 68.3 & 6 & 50.5 & 6 & 72.4 & 5.0 \\
GRAPE (DPO)\cite{GRAPE2024}               & 87.6 & 3 & 91.2 & 4 & 82.2 & 3 & 55.8 & 3 & 79.2 & 3.3 \\
Octo (SFT)\cite{Octo2024}                 & 78.9 & 5 & 85.7 & 6 & 84.6 & 2 & 51.1 & 5 & 75.1 & 4.5 \\
OpenVLA (SFT)\cite{OpenVLA2024}           & 84.7 & 4 & 88.4 & 5 & 79.2 & 5 & 53.7 & 4 & 76.5 & 4.5 \\
VLA-RL \cite{VLA_RL2025}                  & 90.2 & 2 & 91.8 & 3 & 82.2 & 3 & 59.8 & 2 & 81.0 & 2.5 \\
$\pi_0$-FAST\cite{pi0_2024}               & 96.4 & -- & 96.8 & -- & 88.6 & -- & 60.2 & -- & 85.5 & -- \\
\midrule
$\pi_0$-\textbf{FPO (Ours)}               & \textbf{97.2} & \textbf{1} & \textbf{97.3} & \textbf{1} & \textbf{89.4} & \textbf{1} & \textbf{65.3} & \textbf{1} & \textbf{87.2} & \textbf{1} \\
\bottomrule
\end{tabular}
\end{table*}

\begin{figure*}[t]
 \centering
 \includegraphics[width=0.80\textwidth]{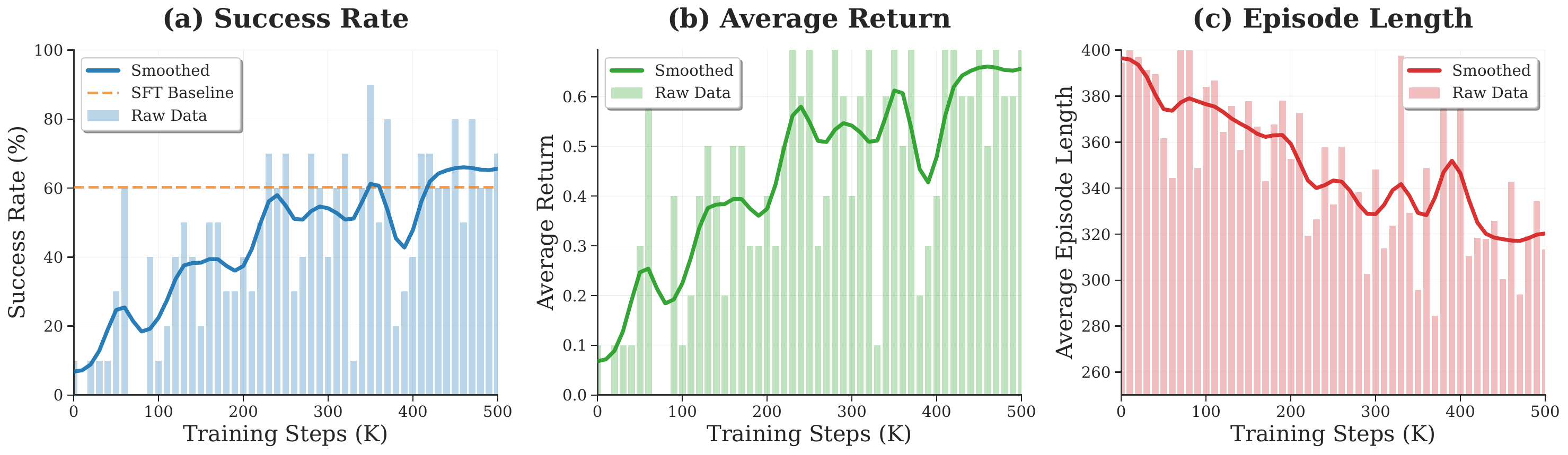} 
 \caption{LIBERO-Long simulation: online fine-tuning curves on a representative task.  }
 \label{fig:libero_90_online_curve}
\end{figure*}

\paragraph{Latent-Space Exploration and Data Handling}
Exploration is induced by multi-step Euler integration in the actor’s latent dynamics. Starting from a sampled latent $x_t^{(0)}\!\sim\!\pi_\theta(\cdot\mid s_t)$ and the CFM velocity field $v_\theta$, we apply $K$ short steps:
\begin{equation}
\resizebox{.85\hsize}{!}{%
  $x_t^{(k+1)} = x_t^{(k)} + \eta\, v_\theta\bigl(x_t^{(k)},\,\tau^{(k)} \mid s_t\bigr), \quad k = 0, \ldots, K - 1$
}
\label{eq:euler_explore}
\end{equation}
where $\{\tau^{(k)}\}$ is a discretization of the flow time and $\eta>0$ is a small step size. The final $x_t^{(K)}$ is decoded by the frozen base. This procedure yields smooth, temporally correlated perturbations that remain aligned with the actor’s generative field. Transitions are stored in a compact sliding-window \emph{trajectory buffer} $\mathcal{B}$ that retains only recent rollouts. Each update draws sample batches from $\mathcal{B}$ and performs several SGD epochs while evicting older entries, which limits distributional drift between the update policy and the data-collecting policy and keeps the loss differential $\Delta\ell_{\text{cfm},t}$ (Eq.~\ref{eq:lossdrop}) evaluated under a distribution close to behavior, stabilizing the ratio mapping in Eq.~\ref{eq:zscore_ratio}.

\section{EXPERIMENTS}

In this section, we empirically evaluate FPO along three axes: (i) final performance on standard manipulation benchmarks relative to strong baselines. (ii) learning dynamics under online interaction (improvement curves and exploration behavior). and (iii) ablations that isolate the contribution of each component.

\subsection{Experimental Setup}

\paragraph{Tasks and Evaluation}
The proposed FPO algorithm was evaluated on two simulated visuomotor benchmarks: LIBERO~\cite{LIBERO2023} and ALOHA Transfer Cube~\cite{ALOHA2024} (as shown in Fig.~\ref{fig:environments}).  LIBERO~\cite{LIBERO2023} comprises four sub-suites—Spatial, Object, Goal, and LIBERO-Long. ALOHA Transfer Cube~\cite{ALOHA2024} is a bimanual manipulation task with contact-rich dynamics. We follow the official success criteria and report per-suite success rate (SR, \%).

\paragraph{Baselines and Protocol}
We compare against $\pi_0$-FAST~\cite{FAST2025}, GRAPE~\cite{GRAPE2024}, Diffusion Policy~\cite{DiffusionPolicy2023}, OpenVLA~\cite{OpenVLA2024}, Octo~\cite{Octo2024}, and VLA-RL~\cite{VLA_RL2025}, covering supervised steering of $\pi_0$, preference alignment, diffusion-based control, large-scale SFT VLAs, and online RL with autoregressive heads. Evaluations follow the official success metrics and protocols. We use public checkpoints when available, otherwise authors’ reference implementations with reported settings. Task definitions, observation/action interfaces, and evaluation seeds are matched across methods. Our runs initialize from the released $\pi_0$ checkpoint, keep the $\pi_0$ decoder frozen, and update only the flow actor and an ensemble critic online.


\subsection{Performance Evaluation and Analysis}

\paragraph{Performance Advantages of FPO on the LIBERO Benchmark}

FPO (denoted as $\pi_0$-\textbf{FPO}) achieves state-of-the-art performance across all four task suites of the LIBERO benchmark, as shown in Table~\ref{tab:libero}. It attains suite-leading success rates with an overall average of 87.2$\%$, outperforming all baseline methods. On the LIBERO-Long suite, FPO attains 65.3\% SR. This corresponds to improvements of +5.5 percentage points over the RL baseline VLA-RL (59.8\%), +9.5 over GRAPE (55.8\%), and +5.1 over $\pi_0$-FAST (60.2\%), as reported in Table~\ref{tab:libero}. These comparisons situate FPO ahead of both online RL and offline-trained baselines under the same evaluation protocol.

This improvement indicates that even foundation models trained on large-scale offline datasets retain unused capacity. FPO leverages an online, reward-driven update mechanism to exploit this capacity, addressing the lack of adaptability and exploration inherent to purely offline training. These results demonstrate that FPO can effectively learn from sparse rewards and refine goal-directed behaviors, extending performance beyond the limits of imitation-based methods. 

\paragraph{FPO Enables Stable and Efficient Online Learning}


To further assess the effectiveness of FPO in improving model performance through online reinforcement learning, we examine its learning dynamics on the LIBERO benchmark and the ALOHA Transfer Cube task. Starting from an SFT baseline, FPO consistently increases both success rate and average return throughout training, as shown in Fig.~\ref{fig:libero_90_online_curve}. The return curve exhibits a similar upward trend, while episode length remains 
steady downward trend, indicating that gains arise from discovering more direct and efficient strategies rather than extending trial duration.

\begin{figure}[!t]
  \centering
  \resizebox{\linewidth}{!}{%
    \includegraphics{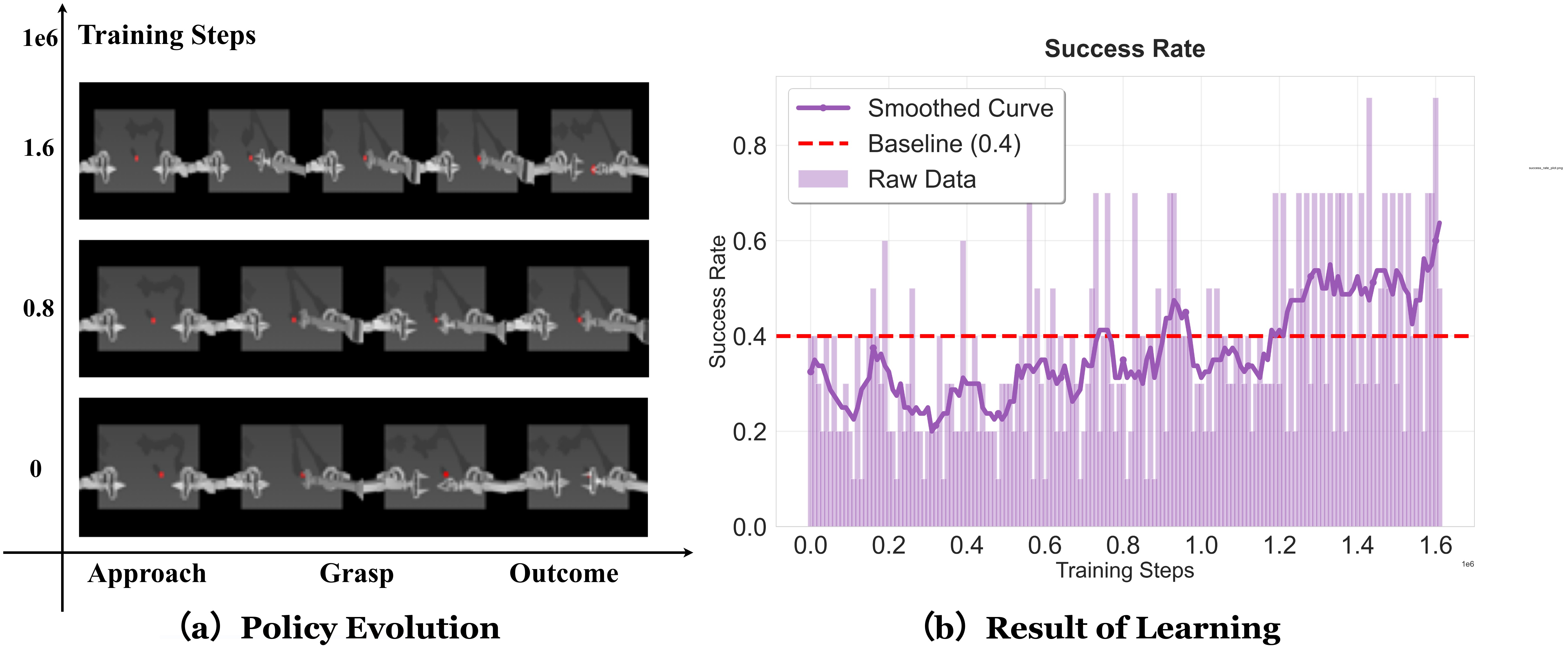}%
  }
  \caption{FPO online learning on the ALOHA Transfer Cube task.
  \textbf{(a)} Policy evolution at 0/0.8M/1.6M training steps: the baseline side-grasp failure mode is corrected to a robust top-down grasp that consistently completes the task.
  \textbf{(b)} Success rate (SR) curve: the smoothed trajectory (purple) steadily improves, surpassing the 40$\%$ baseline (red dashed) and reaching 65$\%$, mirroring the behavioural change in (a).}
  \label{fig:aloha_curves_combined}
\end{figure}

Comparable behavior is observed in the ALOHA Transfer Cube task (Fig.~\ref{fig:aloha_curves_combined}). Beginning with the $\pi_0$ model at $\sim40\%$ success, FPO exceeds 65$\%$ after a comparable training budget. The smoothed trajectory improves monotonically and avoids the instability typically seen in online RL under sparse rewards. Together, these results indicate stable online learning across distinct manipulation domains, supporting FPO as an effective fine-tuning framework for VLA policies. 

\subsection{Analysis of FPO's Internal Learning Mechanism}

\begin{figure}[!ht]
    \centering
    \resizebox{0.85\linewidth}{!}{%
        \includegraphics{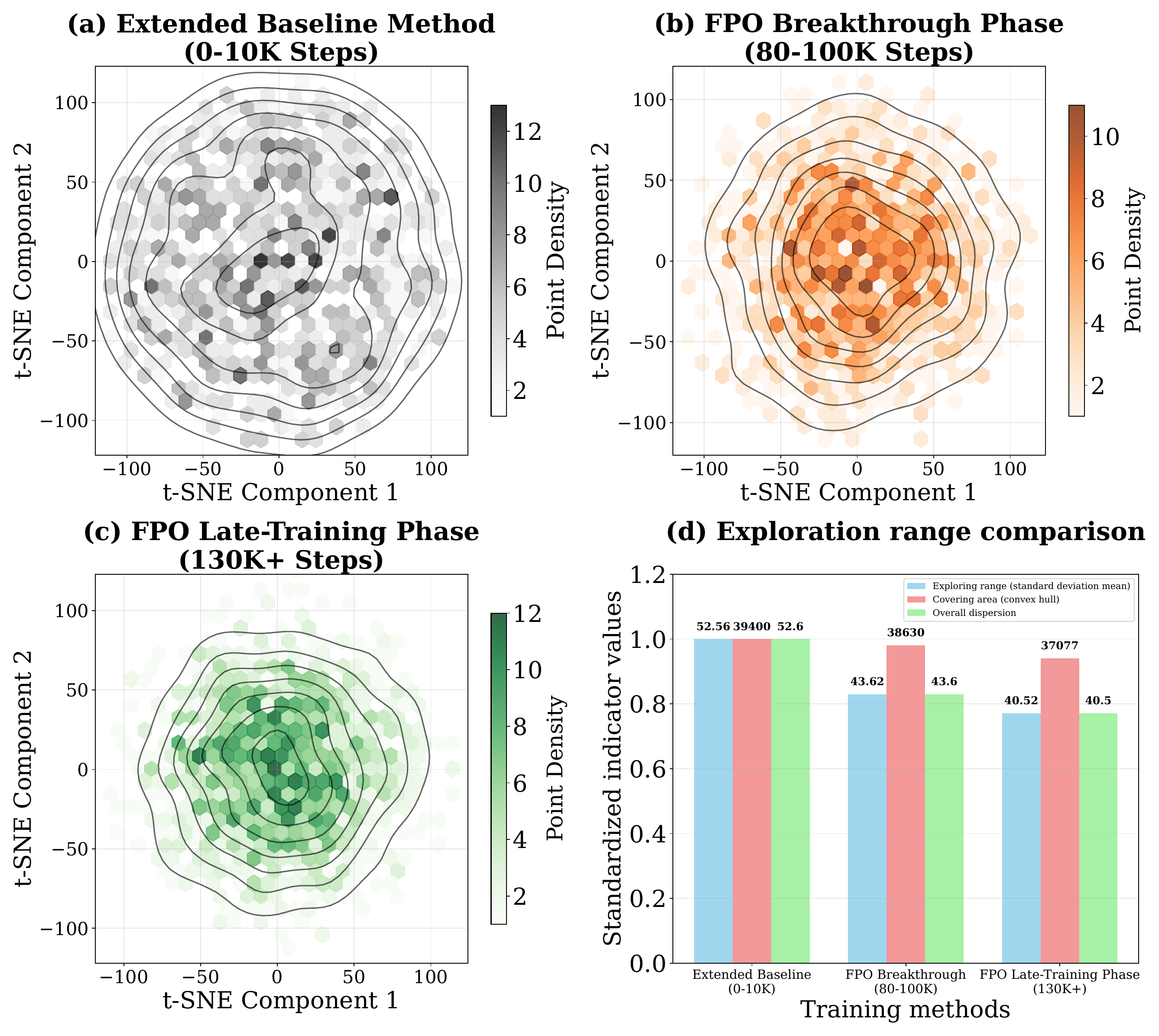}%
    }
    \caption{FPO latent action space evolution. Visualized via t-SNE, this figure shows the policy's latent action distribution transitioning from broad exploration to focused exploitation across training stages. \textbf{(a)} Initial policy: wide, high-variance exploration. \textbf{(b)} Breakthrough phase: distribution concentrates around successful sequences. \textbf{(c)} Late-Training Phase: highly focused, low-variance exploitation of optimal regions. \textbf{(d)} Bar chart: quantifies reduced exploration range and dispersion, confirming convergence to refined behaviors.}
    \label{fig:latent_evolution}
\end{figure}

To analyze how FPO achieves its performance, we examined the evolution of its internal behavior during training by visualizing the distribution of latent action chunks across different stages, shown as Fig.~\ref{fig:latent_evolution}. Using t-SNE for dimensionality reduction, the results reveal a clear trajectory from broad exploration to focused exploitation in the latent space.

In the initial phase (Fig.~\ref{fig:latent_evolution}(a)), the policy, guided by the imitation prior, explores a wide region of the latent space, enabling the discovery of rewarding areas beyond the baseline policy. During the breakthrough phase (Fig.~\ref{fig:latent_evolution}(b)), exploration becomes more structured and concentrated around successful action sequences, indicating prioritization of high-value regions. At convergence period (Fig.~\ref{fig:latent_evolution}(c)), the distribution narrows into a low-variance cluster, reflecting efficient exploitation of the optimal region. The bar chart in Fig.~\ref{fig:latent_evolution}(d) quantifies this transition, showing a marked reduction in dispersion and variance over training. These results demonstrate that FPO supports gradient-driven exploration beyond imitation priors and enables stable convergence to efficient task-solving behaviors.

\begin{figure}[ht]
    \centering
    \includegraphics[width=\columnwidth]{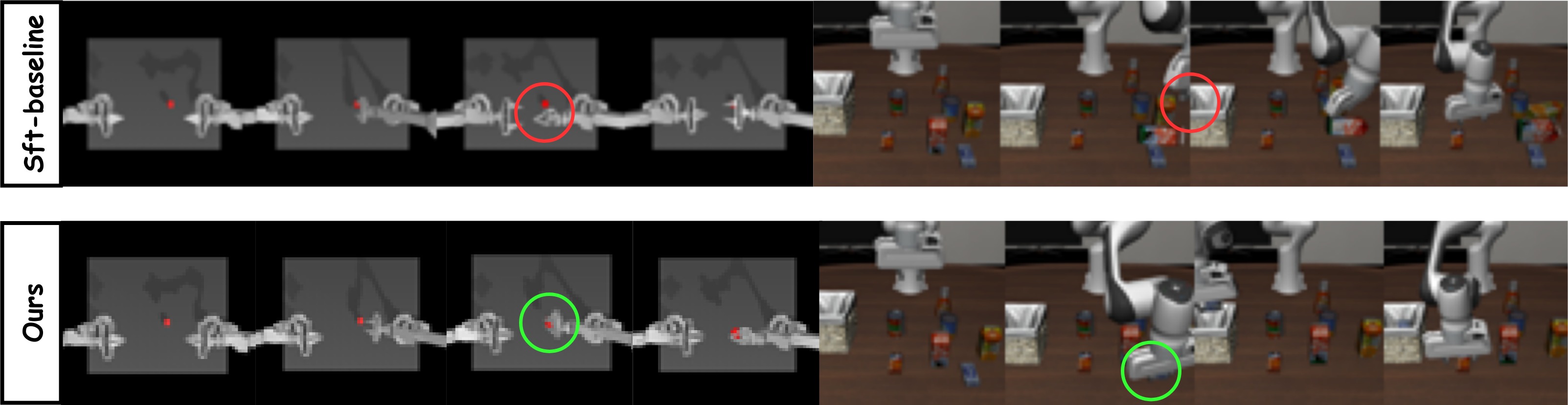}
    \caption{FPO's ability to correct suboptimal behaviors. \textbf{(Top)} The SFT baseline policy consistently fails the task due to a suboptimal grasping approach inherited from the imitation prior. \textbf{(Bottom)} After online fine-tuning with FPO, the policy discovers a novel and successful trajectory from the same initial state, showcasing effective online correction.}
    \label{fig:case_study}
\end{figure}

In addition, the Fig.~\ref{fig:case_study} visually demonstrates FPO's ability to resolve specific, recurring failure modes. The pre-trained $\pi_0$ policy often fails in a representative LIBERO task by attempting a suboptimal side grasp, leading to object instability (Fig.~\ref{fig:case_study}, top). FPO, after online fine-tuning, fundamentally alters this approach, consistently executing a robust top-down grasp from the same initial state that previously led to failure (Fig.~\ref{fig:case_study}, bottom). This illustrates FPO's capacity to learn physically grounded, effective trajectories through active online interaction, fixing nuanced, contact-rich errors that are challenging for offline methods alone.

\subsection{Ablation Studies}

\begin{table}[!t]
    \centering
    \caption{Ablations on LIBERO-90 task \emph{pick up the butter and put it in the tray}.
    Numbers are final success rate (\%). Each row removes one component from FPO.}
    \label{tab:ablation}
    \setlength{\tabcolsep}{5pt}
    \renewcommand{\arraystretch}{1.12}
    \begin{tabular}{@{}p{0.68\linewidth} c@{}}
        \toprule
        Method                                           & Success Rate (\%) \\
        \midrule
        FPO (complete)                                   & 78.5 \\
        \hspace{0.6em}– without CFM ratio proxy          & 32.4 \\
        \hspace{0.6em}– without PPO clipping             & 45.1 \\
        \hspace{0.6em}– single-step integration ($K{=}1$)& 61.7 \\
        \hspace{0.6em}– single critic (without ensemble) & 71.2 \\
        \bottomrule
    \end{tabular}
\end{table}

To assess which components drive FPO’s performance, we conduct ablations on the LIBERO-90 task \emph{pick up the butter and put it in the tray}. Table~\ref{tab:ablation} reports final success rate (\%) after training. Each variant disables exactly one component while keeping the training budget, architectures, and hyperparameters fixed. We consider four interventions: substituting the CFM-based ratio with an SAC-style latent-space update, removing PPO-style clipping, reducing exploration to single-step integration, and replacing the critic ensemble with a single critic. All ablations degrade performance relative to the complete method: replacing the CFM-based ratio causes the most substantial drop, removing clipping also leads to a marked reduction, limiting exploration depth yields a smaller but consistent decline, and using a single critic has the least impact yet remains non-negligible. Taken together, these results indicate that all components are consequential—the structure-aligned ratio and trust-region control account for a large portion of the gains, while exploration depth and value ensembling contribute additional stability and data efficiency.

\paragraph{Importance of the CFM-based Ratio Proxy.}
Replacing the CFM-based policy ratio proxy with standard Soft Actor-Critic (SAC) in the latent space significantly reduced the success rate from 78.5$\%$ to 32.4$\%$. This indicates that FPO’s performance depends critically on the structurally-aware update rule that leverages the generative loss. Conventional RL in latent space is insufficient to achieve comparable results.

\paragraph{Effect of PPO-style Clipping.}
Removing PPO-style clipping caused instability and reduced success to 45.1$\%$ (-33.4$\%$). This confirms the role of clipping as a trust-region mechanism \cite{Schulman2017PPO}, preventing uncontrolled updates from the strong CFM signal and avoiding policy collapse.

\paragraph{Role of Multi-step Exploration.}
Disabling multi-step Euler integration ($K=1$) lowered success to 61.7$\%$ (-16.8$\%$). This shows the benefit of generating temporally correlated latent trajectories, which improve the discovery of viable action sequences and enable stable execution in contact-rich tasks.

\paragraph{Contribution of the Q-Ensemble.}
Using a single critic instead of a Q-ensemble reduced success to 71.2$\%$. Although the effect is smaller (-7.2$\%$), the ensemble improves stability by providing more reliable advantage estimates, which is particularly useful in sparse-reward settings.


\subsection{Latent Space Characteristics of Successful Policies}
\begin{figure}[!ht]
    \centering
    \includegraphics[width=0.85\columnwidth]{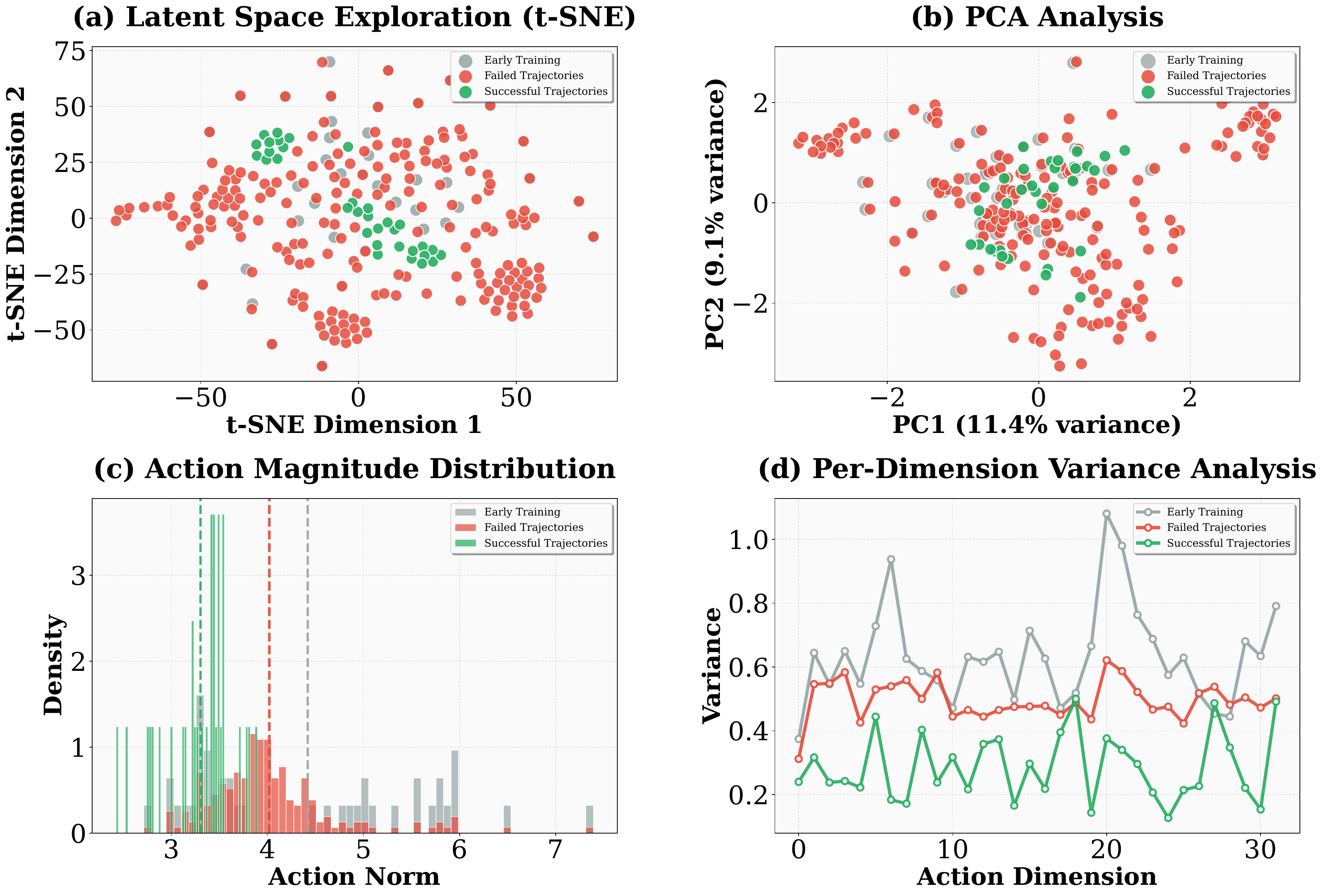}
    \caption{Latent space analysis on initial policy, failed and successful Trajectories. The t-SNE and PCA plots (top row) demonstrate that successful trajectories (green) converge to a distinct, highly structured region of the latent space, sharply contrasting with the diffuse distributions of the initial policy (blue) and failed attempts (red). The action magnitude distribution (bottom-left) shows successful trajectories favoring a narrower, optimal range of action norms. Critically, the per-dimension variance plot (bottom-right) reveals significantly lower variance across most latent dimensions for successful trajectories, indicating learned precision and intentionality.}
    \label{fig:comprehensive_analysis}
\end{figure}

To examine latent space characteristics distinguishing successful from failed policy executions, we conducted a statistical analysis of latent action chunks from the untrained imitation prior, failed rollouts after partial training, and successful rollouts after extended training. Fig.~\ref{fig:comprehensive_analysis} summarizes the statistical differences across these groups. The t\mbox{-}SNE and PCA projections (Fig.~\ref{fig:comprehensive_analysis}a,b) show that successful trajectories occupy a compact and well-defined region of the latent space, whereas actions from the initial policy and failed rollouts are more dispersed. This separation indicates that FPO guides the policy toward a higher-performing latent subspace.

Analysis of action magnitudes (Fig.~\ref{fig:comprehensive_analysis}c) reveals that successful trajectories fall within a narrower range, suggesting avoidance of extreme actions and convergence toward an effective magnitude. The per-dimension variance plot (Fig.~\ref{fig:comprehensive_analysis}d) further shows reduced variance across most dimensions, confirming that successful policies act with greater precision by suppressing unnecessary exploratory noise and focusing capacity on reliable execution. This transition reflects effective skill acquisition in the latent space.

Quantitatively, successful rollouts exhibit higher silhouette scores and larger Mahalanobis distance to the success centroid, alongside lower within-cluster variance and reduced first-order temporal differences, indicating concentration of probability mass in a stable high-value latent region.

\section{CONCLUSION}
This work introduced a FPO algorithm for online fine-tuning of flow-matching VLA policies. FPO resolves the incompatibility with conventional policy-gradient methods by replacing explicit likelihood ratios with a likelihood-free proxy derived from per-sample changes in the conditional flow-matching objective, thereby enabling PPO-style clipped updates without Jacobian or density evaluation. The method incorporates structure-aware credit assignment in the latent space, a clipped surrogate objective, multi-step latent exploration, and a Q-ensemble, enabling stable and efficient optimization in sparse-reward and contact-rich environments. Experiments on the LIBERO benchmark and the ALOHA Transfer Cube task demonstrate that $\pi_0$-FPO consistently outperforms imitation-trained priors and strong baselines, including OpenVLA, Octo, Diffusion Policy, GRAPE, and $\pi_0$-FAST. Ablation experiments and latent space dynamics analysis not only confirm the efficacy of individual FPO components, but also demonstrate enhanced mitigation of recurring failure patterns through qualitative assessment. In the future, we will further enhance the few-shot adaptation capability based on limited online interactions, enabling faster learning and transfer while minimizing additional data requirements.

\section*{ACKNOWLEDGMENT}
This study is supported by the State Key Laboratory of Brain Cognition and Brain-inspired Intelligence Technology (Grant No. JS202401), the funding from Institute of Automation, Chinese Academy of Sciences (Grant No. E411230101), and the National Natural Science Foundation of China (Grant No. 62576341 and No. 32441109).

\end{document}